\documentclass[11pt]{article}

\usepackage[preprint]{acl}
\usepackage{times}
\usepackage{latexsym}
\usepackage[T1]{fontenc}
\usepackage[utf8]{inputenc}
\usepackage{microtype}
\usepackage{inconsolata}
\usepackage{amsmath,amssymb}
\usepackage{graphicx}
\usepackage{booktabs}
\usepackage{array}
\usepackage{multirow}
\usepackage{colortbl}
\usepackage{tikz}
\usetikzlibrary{arrows.meta}

\title{Scaling Looped MoE Language Models under Matched Budgets: Information Flow and Parameter Allocation}
\author{
  Shipeng Bai\textsuperscript{\textdagger}, Jiayu Qin\textsuperscript{\textdagger},
  Hong Liu\textsuperscript{\textdagger}, Xili Wang\textsuperscript{\textdagger}, Jianchao Tan\textsuperscript{*}, Jiacheng Li \\
  \textbf{He Zhang, Yifan Lu, Jiaqi Zhang, Kefeng Zhang, Yuchen Xie, Xunliang Cai} \\[4pt]
  {\normalfont Meituan}
}

\begin{document}
\maketitle
\begingroup
\renewcommand{\thefootnote}{\fnsymbol{footnote}}
\footnotetext[2]{Equal contribution.}
\footnotetext[1]{Corresponding author.}
\makeatletter
\insert\footins{%
  \normalfont\footnotesize
  \interlinepenalty\interfootnotelinepenalty
  \splittopskip\footnotesep
  \splitmaxdepth \dp\strutbox \floatingpenalty \@MM
  \hsize\columnwidth \@parboxrestore
  \raggedright
  \parindent 1em\noindent
  \hb@xt@1.8em{\hss}%
  Contact: \href{mailto:baishipeng02@meituan.com,tanjianchao02@meituan.com}{{\fontsize{7.8}{9.5}\selectfont\texttt{\{baishipeng02, tanjianchao02\}@meituan.com}}}.\par
}
\makeatother
\endgroup
\setcounter{footnote}{0}
\suppressfloats[t]
\begin{abstract}
Looped Transformers decouple computational depth from parameter count, and prior studies have demonstrated promising performance gains with increased computation. However, systematic design principles for configuring these models under strict iso-parameter and iso-FLOP constraints remain scarce.
We contribute to this direction by improving information flow in looped models and systematically studying loop parameter configurations. Specifically, we refine this information flow via gated inner-loop state updates and a cross-stage residual connection. 
Building on this design, controlled studies identify a configuration combining a single-pass Core depth of 25\%--35\% of the total unrolled depth with reallocating saved parameters into extra routed experts per layer, improving performance at zero extra compute while preserving compatibility with existing training infrastructure.
Evaluated after 1T pre-training tokens, our 69B MLA and 51B KDA Mixture-of-Experts models outperform budget-matched non-recurrent baselines by an average of $1.16$ and $2.29$ percentage points across four benchmarks, respectively. 
Further reproduction experiments on a common backbone with a 200B-token training budget show that our method achieves lower perplexity than other looped methods. Code for these reproduction experiments is available at \href{https://github.com/lxi01766/Scaling-Looped-MoE-Language-Models-under-Matched-Budgets-Information-Flow-and-Parameter-Allocation}{GitHub}.
\end{abstract}

\begin{figure}[t]
    \centering
    \includegraphics[width=\columnwidth]{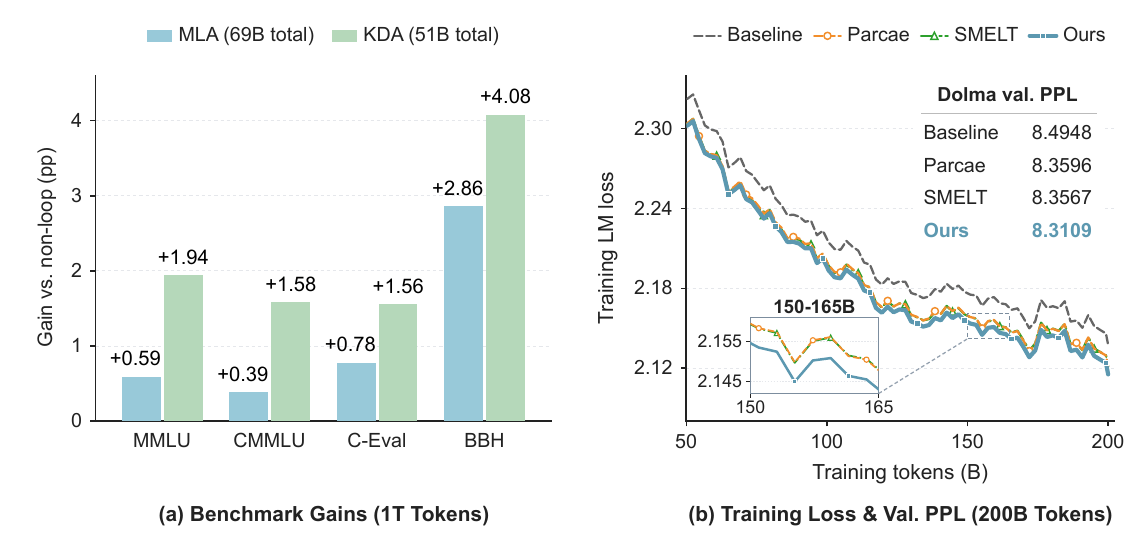}
    \caption{Main results and comparison with existing looped models. (a) Benchmark gains in percentage points (pp) over corresponding non-loop baselines after 1T tokens with MLA or KDA attention. (b) Training LM loss and final validation PPL on Dolma~3 Mix with a common GQA-MoE backbone, trained for 200B tokens.}
    \label{fig:results-overview}
\end{figure}

\section{Introduction}
\label{sec:introduction}

Looped Transformers decouple computational depth from parameter scale through the iterative reuse of weight-shared layers, 
presenting an alternative depth-scaling paradigm orthogonal to conventional parameter expansion~\citep{geiping2025recurrent,prairie2026parcae}.
Prior work demonstrates that such architectures can match or even surpass competitive non-recurrent models on complex reasoning and algorithmic induction tasks.
Recent studies~\citep{wang2026smelt,chen2026loopmoe} have further substantiated the advantages of recurrent computation under strictly matched parameter budgets as well as training and 
serving FLOPs.

Although prior studies have demonstrated the promise of looped architectures, they offer limited guidance on how to scale them systematically under matched parameter and compute budgets. 
Reusing weights across iterations requires balancing input preservation against progressive state updates, whereas varying the recurrent span and iteration count shifts the trade-off 
between effective computational depth and unique parameter capacity---inherently coupling loop topology with budget allocation. These structural dependencies motivate two central research questions:
\textbf{Information Flow within the Loop:} Under weight sharing, how should inner-loop state updates and cross-stage representation transfer be structured to enable effective iterative refinement?
\textbf{Loop Configuration and Cross-Architecture Generality:}
Under iso-parameter and iso-FLOP constraints, how should the looped structure be partitioned with appropriate parameter allocation, 
and does the same recurrent design improve performance across different backbone architectures?

Existing looped models predominantly organize recurrent computation into a three-stage \textit{Prelude--Core--Coda} topology~\citep{geiping2025recurrent,prairie2026parcae}, iteratively updating hidden states by combining them with input representations.
To address the first question, we refine internal information flow through \textit{inner-loop state updates} and a \textit{cross-stage residual connection}.
Firstly, for inner-loop state updates, we stabilize feature scales across loop iterations via boundary normalization and combine historical states with injected inputs through independently controlled gated fusion.
Secondly, we add a cross-stage residual connection linking the Prelude directly to the Coda to restore the direct residual pathway and gradient flow interrupted by boundary normalization and gated fusion.
Together, these simple yet effective designs establish a stable recurrent flow that reconciles low-level feature fidelity with iterative semantic evolution, yielding substantial performance gains.

Beyond information flow within the loop, systematically scaling recurrent models requires principled loop configuration and cross-architecture generality. While recent studies
have begun extending looped architectures to larger-scale models---such as SMELT \citep{wang2026smelt}, which evaluates looping the middle half of a sparse
MoE twice under matched budgets, and SenseNova-Loop~\citep{sensenova2026loop}, which integrates recurrence with hybrid attention and Manifold-Constrained Hyper-Connections (mHC)~\citep{xie2026mhc}---existing efforts
remain restricted to isolated topologies or fixed configurations, leaving generalizable allocation principles largely unexplored. To bridge this gap, we
systematically investigate the joint design of loop topology and parameter capacity under strictly matched budgets, i.e., iso-parameter and iso-FLOP constraints.
Our controlled studies reveal that effective scaling hinges on a structural trade-off between \textit{shared computational depth} and
\textit{independent parameter capacity}: core partitioning and capacity restoration must be jointly determined.
Guided by this insight, we formulate two transferable design principles: (1) \textbf{Recurrent Core Configuration:} Allocating $25\%\text{--}35\%$ of the total unrolled depth 
to the single-pass depth of the shared Core balances computation reuse against independent boundary-layer capacity; and (2) \textbf{Parameter-to-Expert Reallocation:} Transferring weight-sharing parameter savings into extra routed experts per layer improves performance at zero compute cost, offering a better quality--compatibility trade-off than alternative restoration strategies that alter layer or module dimensions.

To evaluate architectural generality, we apply this design to Mixture-of-Experts (MoE) models equipped with Multi-Head Latent Attention (MLA)~\citep{deepseek2024v2} and Kimi Delta Attention (KDA)~\citep{kimi2025linear}. After 1T pre-training tokens, the 69B MLA and 51B KDA looped models outperform budget-matched non-recurrent baselines across four downstream benchmarks, delivering average gains of +1.16 and +2.29 percentage points, respectively. Furthermore, on a common Qwen3-style MoE backbone~\citep{yang2025qwen3} trained for 200B tokens, our method achieves lower perplexity than both Parcae and SMELT. Figure~\ref{fig:results-overview} summarizes these benchmark gains and the separate 200B-token training-loss comparison with Parcae and SMELT.

\begin{figure*}[!t]
    \centering
    \includegraphics[width=\textwidth]{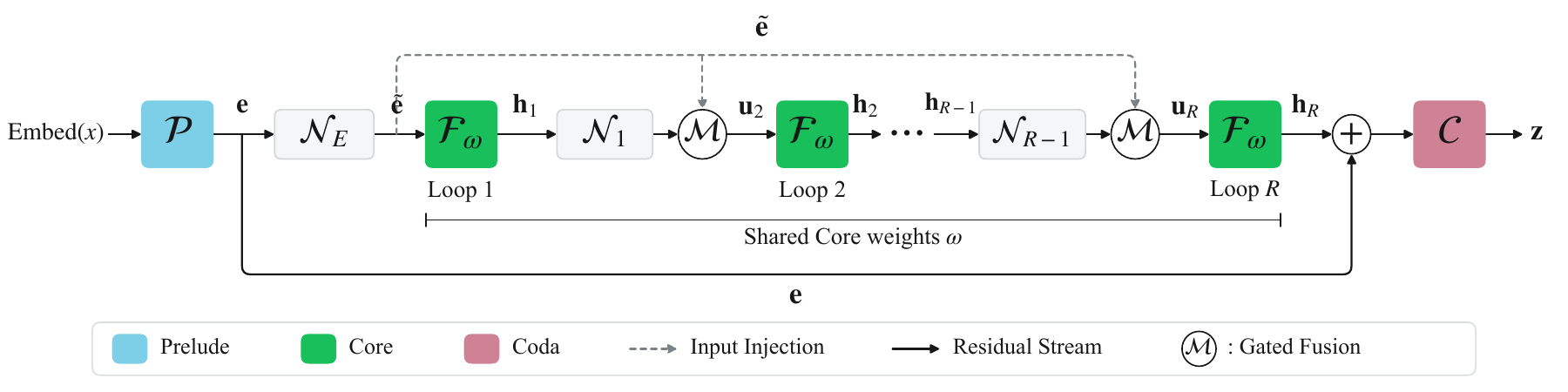}
    \caption{Overview of our looped model with $R$ Core passes. $\mathcal{P}$, $\mathcal{F}_\omega$, and $\mathcal{C}$ denote the Prelude, shared Core, and Coda; all Core passes share weights $\omega$. Given token embeddings $\mathrm{Embed}(x)$, the first pass receives $\tilde{\mathbf{e}}=\mathcal{N}_E(\mathbf{e})$ directly. For $t\geq 2$, boundary normalization $\mathcal{N}_{t-1}$ precedes the gated fusion operator $\mathcal{M}$, which combines the previous state with the fixed input $\tilde{\mathbf{e}}$ (dashed path; Section~\ref{sec:theta_parametrization}). The cross-stage residual connection adds the raw Prelude output $\mathbf{e}$ to $\mathbf{h}_R$ before $\mathcal{C}$ produces $\mathbf{z}$. Output-head operations are omitted.}
    \label{fig:method-overview}
\end{figure*}

\section{Related Work}
\label{sec:related-work}

\paragraph{Looped Language Models and Recurrent Information Flow.}
Early studies explored depth-wise parameter reuse through recurrent computation and cross-layer weight
tying~\citep{dehghani2019universal,lan2020albert}. Subsequent theoretical and empirical work connected
recurrent depth to iterative reasoning and large-scale language modeling~\citep{saunshi2025latent,geiping2025recurrent,zhu2025ouro},
while complementary approaches introduced recurrence into pretrained models~\citep{bae2025recursive,mcleish2025retrofit}.
Recent work further investigates training stability, token-dependent depth allocation, and budget-conditioned
recurrent trajectories~\citep{prairie2026parcae,bae2025recursions,jeddi2026loopformer}.
Building on these developments, we study inner-loop state updates and a cross-stage residual connection,
combining gated fusion with independently parameterized state retention and input injection and a Prelude-to-Coda residual path.

\paragraph{Loop Configuration, Parameter Allocation, and Backbone Applicability.}
Combining recurrence with sparse experts provides flexibility in balancing parameter capacity and
computation~\citep{csordas2024moeut,gao2026loopie}. SMELT~\citep{wang2026smelt} investigates loop topology
through depth--width adjustments and expert expansion under matched budgets, while
LoopMoE~\citep{chen2026loopmoe} emphasizes attention--FFN capacity balancing.
Meanwhile, advances in latent attention, gated linear attention, and residual connectivity broaden the
architectural settings for recurrent computation~\citep{deepseek2024v2,kimi2025linear}. In particular, mHC~\citep{xie2026mhc}
and Attention Residuals~\citep{kimi2026attnres} enrich inter-layer residual information flow, and
SenseNova-Loop~\citep{sensenova2026loop} combines recurrence with hybrid attention and mHC.
Our work investigates how parameter restoration changes the preferred Core partition and compares allocation
strategies under iso-parameter and iso-FLOP budgets. We further evaluate the same recurrent formulation across
MLA- and KDA-based MoE models, alongside a separate study of its compatibility with mHC.

\section{Information Flow Design for Looped Transformers}
\label{sec:method}

\subsection{Base Architecture}
\label{sec:architecture}
We build on the three-stage \textit{Prelude--Core--Coda} architecture~\citep{geiping2025recurrent,prairie2026parcae}, where the \textit{Prelude} $\mathcal{P}$ extracts initial representations, the weight-shared \textit{Core} $\mathcal{F}_\omega$ iteratively refines hidden states over $R$ passes, and the \textit{Coda} $\mathcal{C}$ maps the final state to next-token predictions (Figure~\ref{fig:method-overview}).
Given a token sequence $x$ of length $n$ and hidden dimension $d$, a standard three-stage forward pass computes:
\begin{equation}
    \begin{aligned}
        \mathbf{e} &= \mathcal{P}(\mathrm{Embed}(x)), \\
        \mathbf{h}_t &= \mathcal{F}_\omega\!\big(\mathcal{M}(\mathbf{h}_{t-1}, \mathbf{e})\big), \quad t = 1, \dots, R, \\
        \mathbf{z} &= \mathcal{C}(\mathbf{h}_R),
    \end{aligned}
    \label{eq:base-arch}
\end{equation}
where $\mathbf{e} \in \mathbb{R}^{n \times d}$ is the Prelude output, $\mathbf{h}_t$ is the hidden state after the $t$-th Core iteration, and $\mathcal{M}$ denotes the fusion operator combining the previous state with $\mathbf{e}$.
Throughout the forward pass, $\mathcal{P}$ and $\mathcal{C}$ execute once, whereas $\mathcal{F}_\omega$ reuses parameters $\omega$ across all $R$ passes.
Below, we refine this framework along two axes: inner-loop state updates within the Core (Section~\ref{sec:theta_parametrization}) and a cross-stage residual connection from the Prelude to the Coda (Section~\ref{sec:long-residual}).

\subsection{Inner-Loop State Updates}
\label{sec:theta_parametrization}

Within the weight-shared Core, iterative refinement requires coordinating the fixed input representation with evolving hidden states across passes.
We formulate inner-loop state updates through a direct first-iteration pass, boundary normalization, and a channel-wise gated fusion operator $\mathcal{M}$ (Figure~\ref{fig:method-overview}).

\paragraph{Direct First-Iteration Pass and Gated State Transition.}
Given the Prelude output $\mathbf{e}$, we apply Root Mean Square Layer Normalization~\citep{zhang2019rmsnorm} $\mathcal{N}_E$ to obtain $\tilde{\mathbf{e}} = \mathcal{N}_E(\mathbf{e})$ and route it directly into the shared Core for the first pass:
\begin{equation}
    \mathbf{h}_1 = \mathcal{F}_\omega(\tilde{\mathbf{e}}), \qquad \tilde{\mathbf{e}} = \mathcal{N}_E(\mathbf{e}).
    \label{eq:first-iter}
\end{equation}
Across repeated passes through $\mathcal{F}_\omega$, unnormalized residual updates accumulate and cause the hidden-state magnitude to grow rapidly relative to the fixed input representation, which can overwhelm the injected input signal.
For subsequent iterations $t \in \{2, \dots, R\}$, we therefore apply inter-step boundary RMSNorm $\mathcal{N}_{t-1}$ to align the scale of the recurrent state with the normalized input $\tilde{\mathbf{e}}$ prior to fusion:
\begin{equation}
    \begin{aligned}
        \mathbf{u}_t &= \mathcal{M}\!\big(\mathcal{N}_{t-1}(\mathbf{h}_{t-1}),\, \tilde{\mathbf{e}}\big) \\
        &= \alpha \odot \mathcal{N}_{t-1}(\mathbf{h}_{t-1}) + \Delta \odot B(\tilde{\mathbf{e}}), \\
        \mathbf{h}_t &= \mathcal{F}_\omega(\mathbf{u}_t),
    \end{aligned}
    \label{eq:gated-transition}
\end{equation}
where $B \in \mathbb{R}^{d \times d}$ is a learnable projection matrix and $\odot$ denotes the Hadamard product.
The channel-wise state retention coefficient $\alpha \in (0, 1)^d$ and input injection scale $\Delta \in \mathbb{R}_{>0}^d$ are independently parameterized by separate learnable vectors $\theta, b \in \mathbb{R}^d$:
\begin{equation}
    \alpha = \exp(-\exp(\theta)), \qquad \Delta = \mathrm{softplus}(b).
    \label{eq:independent-coeffs}
\end{equation}
Here, the Core weights $\omega$ are shared across all $R$ passes, $\{B, \theta, b\}$ are tied across iterations $t \ge 2$, and each boundary normalization $\mathcal{N}_{t-1}$ uses iteration-specific scale parameters strictly confined to internal gated fusion.

\paragraph{Relation to Prior Gated Fusion.}
Prior recurrent designs such as Huginn~\citep{geiping2025recurrent} and Parcae~\citep{prairie2026parcae} initialize $\mathbf{h}_0$ with isotropic Gaussian noise and inject inputs at every step; in particular, Parcae adopts channel-wise gated fusion with $\alpha = \exp(-\Delta \odot \exp(a))$ and $\Delta = \mathrm{softplus}(b)$.
Our formulation differs in two key respects:
\begin{itemize}
    \item \textbf{Direct first-iteration input.} Initializing $\mathbf{h}_0$ with unconditioned Gaussian noise forces a shared retention coefficient $\alpha$ to simultaneously suppress stochastic noise at $t=1$ and preserve structured semantic trajectories at $t \ge 2$. Anchoring the first pass directly on $\tilde{\mathbf{e}}$ in Eq.~\eqref{eq:first-iter} eliminates this optimization conflict so that recurrent feedback operates exclusively on meaningful Core states.
    \item \textbf{Independent retention and injection.} In Parcae's parameterization $\alpha = \exp(-\Delta \odot \exp(a))$, state retention and input injection are coupled through $b$: holding $a$ fixed yields
    \begin{equation}
        \frac{\partial \alpha_i}{\partial \Delta_i} = -\alpha_i \exp(a_i) < 0, \qquad i = 1, \dots, d,
        \label{eq:parcae-coupling}
    \end{equation}
    so increasing the input injection scale $\Delta_i$ inevitably suppresses state retention $\alpha_i$ for a fixed $a_i$ (creating gradient coupling through $b_i$).
    By contrast, our formulation in Eq.~\eqref{eq:independent-coeffs} parameterizes $\alpha$ and $\Delta$ with separate vectors $\theta$ and $b$ (corresponding to a bijective channel-wise mapping $\theta = a + \log \Delta$), preserving the exact functional range $\alpha \in (0, 1)$ while ensuring $\partial \alpha / \partial b = 0$.
    State retention and input injection thus remain jointly optimized under the training objective without direct parameter entanglement.
\end{itemize}

\subsection{Cross-Stage Residual Connection}
\label{sec:long-residual}

In a standard Pre-Norm Transformer without loop boundaries, an unobstructed direct residual pathway connects early representations to later layers.
In the recurrent architecture above, however, inserting $\mathcal{N}_E$ before the first Core pass and boundary normalization $\mathcal{N}_{t-1}$ (together with gated fusion $\mathcal{M}$) between iterations---while essential for aligning feature scales and preventing growing recurrent states from overwhelming $\tilde{\mathbf{e}}$---interrupts this direct residual pathway from the Prelude to the Coda.
Breaking this shortcut restricts direct gradient propagation back to the Prelude and forces the terminal state $\mathbf{h}_R$ to simultaneously encode multi-step refinements and preserve initial low-level representations.

To restore a direct residual pathway without compromising internal scale alignment, we introduce a \textbf{cross-stage residual connection} that bypasses $\mathcal{N}_E$, the shared Core, and all boundary normalization layers $\mathcal{N}_{t-1}$.
Specifically, we add the unnormalized Prelude output $\mathbf{e}$ directly to the final recurrent state $\mathbf{h}_R$ before entering the Coda:
\begin{equation}
    \mathbf{z} = \mathcal{C}(\mathbf{e} + \mathbf{h}_R).
    \label{eq:long-residual}
\end{equation}
This single cross-stage addition introduces no extra trainable parameters, preserves an unobstructed gradient path from the Coda to the Prelude, and allows the recurrent Core to focus on learning complementary iterative refinements on top of the initial representation.

\section{Loop Configuration and Parameter Allocation}
\label{sec:loop-configuration}
\label{sec:budget}

Building upon the information-flow design in Section~\ref{sec:method}, we systematically examine the joint
configuration of loop topology and parameter capacity. At a fixed unrolled depth and per-layer
configuration, changing the loop partition alters the number of independently parameterized layers
and, consequently, the capacity that must be restored to meet the parameter budget. We therefore
study loop configuration and budget restoration jointly: first investigating recurrence count and
Core depth under iso-parameter and iso-FLOP budgets, then comparing restoration strategies and
evaluating the effect of parameter restoration.

\subsection{Recurrent Core Configuration}
\label{sec:recurrent-core-configuration}

\begin{figure}[t]
    \centering
    \includegraphics[width=\columnwidth]{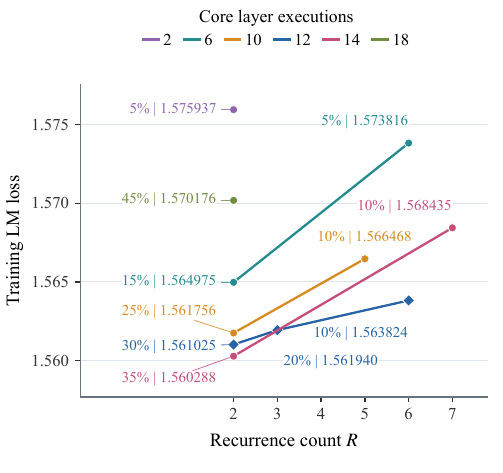}
    \caption{Training LM loss after 170B tokens under iso-parameter and iso-FLOP budgets, with 20 unrolled
    layers and parameter capacity restored via uniform expert-pool expansion (Section~\ref{sec:parameter-restoration}). Colors group configurations with the same total number of Core layer executions and fixed
    Prelude and Coda depths. Point labels give the Core depth ratio $\rho$ as a percentage, followed by
    loss. The run with three iterations and 18 Core layer executions diverged (loss $6.931539$) and is omitted
    from the plotted range.}
    \label{fig:recurrence-count-choice}
\end{figure}

We examine recurrent Core configuration along two dimensions: recurrence count and Core depth.
Let $L_{\mathrm{pre}}$, $L_{\mathrm{core}}$, and $L_{\mathrm{coda}}$ denote the numbers of layers in the
Prelude, Core, and Coda, respectively, and let $R$ denote the recurrence count. We define the total
unrolled depth and the Core depth ratio as
\begin{equation}
    \begin{aligned}
        D_{\mathrm{total}} &= L_{\mathrm{pre}}+R L_{\mathrm{core}}+L_{\mathrm{coda}},\\[-1pt]
        \rho &= \frac{L_{\mathrm{core}}}{D_{\mathrm{total}}}.
    \end{aligned}
    \label{eq:core-depth-ratio}
\end{equation}
The Core depth ratio $\rho$ measures the single-pass depth of the shared Core as a fraction of the
total unrolled depth and corresponds to the percentages annotated in Figures~\ref{fig:recurrence-count-choice} and~\ref{fig:core-partition-restoration}.
We conduct controlled sweeps on an MLA-MoE backbone with approximately 0.8B active and 12.9B total
parameters (including 4.23B over-embedding parameters), fixing $D_{\mathrm{total}}=20$ under
iso-parameter and iso-FLOP budgets via uniform expert-pool restoration (Section~\ref{sec:parameter-restoration}) and reporting training LM loss
after 170B tokens. For example, $\rho=5\%$ and $\rho=35\%$ correspond
to Core depths of one and seven layers, respectively.

\paragraph{Choice of recurrence count.}
We first compare recurrence counts while holding the Prelude and Coda depths fixed. Configurations
connected by the same color in Figure~\ref{fig:recurrence-count-choice} share these boundary-stage depths;
Core depth and recurrence count are jointly adjusted to preserve the total unrolled depth. Two iterations
yield lower training loss than configurations with more iterations across the evaluated comparisons.
For example, with three layers each in the Prelude and Coda, increasing the recurrence count from two
to seven reduces the Core depth ratio from $35\%$ to $10\%$ and raises loss from $1.5603$ to $1.5684$.
We therefore adopt $R=2$ as the default recurrence count, consistent with the two-pass configurations
used in SMELT~\citep{wang2026smelt} and SenseNova-Loop~\citep{sensenova2026loop}.

\paragraph{Core depth allocation.}
Having selected the recurrence count, we compare the two-pass configurations with different Core
depth ratios in Figure~\ref{fig:recurrence-count-choice}. Increasing the ratio from $5\%$ to $25\%$
and $35\%$ reduces training loss from $1.5759$ to $1.5618$ and $1.5603$, respectively. Further
increasing it to $45\%$ raises loss to $1.5702$. At fixed unrolled depth, expanding the shared Core
reduces the number of independent boundary layers: ratios of $25\%$--$35\%$ retain three to five layers
each in the Prelude and Coda, whereas a ratio of $45\%$ leaves only one layer in each stage. These
results show that extending the shared span does not yield monotonic improvements. We consequently
use two Core iterations and allocate $25\%$--$35\%$ of the total unrolled depth to the single-pass
Core in subsequent experiments.

\subsection{Parameter Budget Restoration Strategies}
\label{sec:parameter-restoration}

To compensate for the independent parameter capacity removed by weight sharing, existing approaches
adopt different budget-restoration strategies. SMELT~\citep{wang2026smelt} jointly adjusts model depth
and width, narrowing the hidden dimension to accommodate additional recurrent computation and
increasing the expert count to restore parameter capacity. LoopMoE~\citep{chen2026loopmoe} instead
expands the MLA Q/KV low-rank projections, reduces individual expert widths, and enlarges the expert
pool, redistributing capacity between attention and FFN modules.

We incorporate these restoration strategies into our recurrent framework and compare three variants
on an MLA-MoE backbone: depth--width reallocation (Depth--Width), attention--FFN reallocation
(Attn--FFN), and uniform layer-wise expert-pool expansion (Ours). Our strategy restores the parameter
budget by increasing expert counts across physical layers while retaining the hidden dimension,
attention configuration, per-expert FFN width, and number of activated experts. All variants use the
same training data, tokenizer, and learning-rate schedule, with iso-parameter and iso-FLOP budget
targets, and are evaluated at approximately 100B training tokens.

\begin{table}[!htbp]
    \centering
    \definecolor{restorationOursBlue}{HTML}{E5F3FF}
    \fontsize{9}{10.5}\selectfont
    \setlength{\tabcolsep}{3pt}
    \renewcommand{\arraystretch}{1.12}
    \begin{tabular}{@{}>{\columncolor{white}[0pt][\tabcolsep]\raggedright\arraybackslash}p{\dimexpr\columnwidth-135pt-6\tabcolsep\relax}
        *{2}{>{\centering\arraybackslash}p{45pt}}
        >{\columncolor{white}[\tabcolsep][0pt]\centering\arraybackslash}p{45pt}@{}}
        \toprule
        Strategy & $\Delta$ Val. Avg. & $\Delta$ MMLU & $\Delta$ CMMLU \\
        \midrule
        \rowcolor{restorationOursBlue}
        \textbf{Ours} & 0.0000 & 0.0000 & 0.0000 \\
        Depth--Width & $-0.0007$ & $+0.0025$ & $-0.0198$ \\
        Attn--FFN & $+0.0112$ & $+0.0323$ & $+0.0217$ \\
        \midrule
        Strategy & $\Delta$ GSM8K & \shortstack{$\Delta$ FullStack\\Bench} & $\Delta$ OWT2 \\
        \midrule
        \rowcolor{restorationOursBlue}
        \textbf{Ours} & 0.0000 & 0.0000 & 0.0000 \\
        Depth--Width & $+0.0019$ & $+0.0141$ & $-0.0009$ \\
        Attn--FFN & $+0.0086$ & $+0.0181$ & $+0.0090$ \\
        \bottomrule
    \end{tabular}
    \caption{Changes in validation loss relative to Ours. Negative values indicate improvements, while positive
    values indicate degradation. $\Delta$ Val. Avg. denotes the change in the unweighted mean loss across our
    20 held-out validation sets, with five representative sets shown individually.}
    \label{tab:parameter-restoration}
\end{table}

As shown in Table~\ref{tab:parameter-restoration}, Attn--FFN increases mean validation loss by
approximately $0.0112$ relative to Ours, whereas Depth--Width reduces it by approximately $0.0007$.
The aggregate gap between Ours and Depth--Width is small, while their task-level results reveal
different trade-offs: Ours achieves lower losses on MMLU~\citep{hendrycks2021mmlu}, GSM8K~\citep{cobbe2021gsm8k}, and FullStackBench~\citep{liu2024fullstackbench}, whereas
Depth--Width performs better on CMMLU~\citep{li2024cmmlu} and OWT2~\citep{gao2021pile}. Given the small aggregate loss gap and task-dependent
trade-offs, we also consider compatibility with existing implementations. Expert-pool expansion
preserves the original attention and per-expert FFN dimensions, facilitating reuse of existing kernel
implementations. We therefore adopt uniform layer-wise expert-pool expansion as the default
restoration strategy.

\begin{figure}[!t]
    \centering
    \includegraphics[width=\columnwidth]{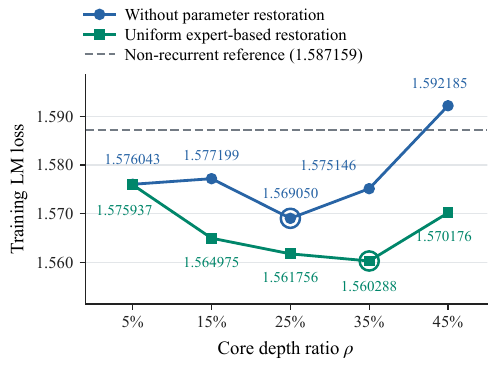}
    \caption{Effect of parameter restoration across Core depth ratios $\rho$. All models have 20 unrolled
    layers and two Core iterations ($R=2$). Values are training LM loss after 170B tokens. Green: uniform
    layer-wise expert expansion; blue: no restoration. The dashed line marks the non-recurrent
    baseline; circles identify each curve's best configuration.}
    \label{fig:core-partition-restoration}
\end{figure}

\paragraph{Effect of parameter restoration.}
We further evaluate the selected strategy by comparing Core configurations with and without parameter
restoration. These experiments use two Core iterations and 20 unrolled layers, retain the same loop
topology before and after restoration, and report training LM loss after 170B tokens. As shown in
Figure~\ref{fig:core-partition-restoration}, uniform layer-wise expert expansion lowers loss across all
evaluated configurations. For Core depth ratios of $25\%$ and $35\%$, restoration reduces loss from
$1.5691$ and $1.5751$ to $1.5618$ and $1.5603$, respectively. These results demonstrate that
replenishing the parameter capacity removed by weight sharing through expert-pool expansion further
improves training loss.

\definecolor{expOursBlue}{HTML}{E5F3FF}
\definecolor{expGainInk}{HTML}{465C70}
\newcommand{\expresult}[2]{\textbf{#1}\,{\footnotesize\textcolor{expGainInk}{(+#2)}}}
\newlength{\expScoreCol}

\begin{table*}[!t]
    \centering
    \small
    \setlength{\tabcolsep}{4pt}
    \setlength{\expScoreCol}{\dimexpr(\linewidth-80pt-12\tabcolsep)/5\relax}
    \renewcommand{\arraystretch}{1.0}
    \setlength{\aboverulesep}{2.5pt}
    \setlength{\belowrulesep}{2.5pt}
    \begin{tabular}{@{}p{43pt}p{37pt}*{4}{>{\centering\arraybackslash}p{\expScoreCol}}>{\columncolor{white}[\tabcolsep][0pt]\centering\arraybackslash}p{\expScoreCol}@{}}
        \toprule[0.9pt]
        \textbf{Attention} & \textbf{Model}
        & MMLU & CMMLU & C-Eval & BBH & \textbf{Avg.} \\
        \midrule[0.45pt]
        & Baseline & 70.44 & 75.96 & 76.31 & 48.09 & 67.70 \\
        \rowcolor{expOursBlue}
        \cellcolor{white}\multirow{-2}{*}{\textbf{MLA}} & \textbf{Ours}
        & \expresult{71.03}{0.59} & \expresult{76.35}{0.39} & \expresult{77.09}{0.78}
        & \expresult{50.95}{2.86} & \expresult{68.86}{1.16} \\
        \midrule[0.45pt]
        & Baseline & 68.69 & 74.92 & 73.83 & 44.24 & 65.42 \\
        \rowcolor{expOursBlue}
        \cellcolor{white}\multirow{-2}{*}{\textbf{KDA}} & \textbf{Ours}
        & \expresult{70.63}{1.94} & \expresult{76.50}{1.58} & \expresult{75.39}{1.56}
        & \expresult{48.32}{4.08} & \expresult{67.71}{2.29} \\
        \bottomrule[0.9pt]
    \end{tabular}
    \caption{Main results at 1T training tokens. Scores are percentages (higher is better);
    parentheses show gains in pp. Avg. is the unweighted mean of all four benchmarks.}
    \label{tab:main-results}
\end{table*}

Together, these findings motivate a unified configuration with two Core iterations, a single-pass
Core depth of $25\%$--$35\%$ of the total unrolled depth, and uniform layer-wise expert expansion.
We evaluate this configuration across distinct attention backbones in the next section.

\section{Scaling Validation and Ablations}
\label{sec:experiments}

Our experiments examine \textit{inner-loop state updates and the cross-stage residual connection},
together with the \textit{applicability of the unified recurrent design across modern backbones}.
We evaluate large-scale pre-training with MLA and KDA attention under iso-parameter and iso-FLOP budgets,
compare against Parcae and SMELT on a common backbone using public Dolma~3 Mix data~\citep{olmo2025olmo3}, and analyze components through
controlled ablations. Experiments with mHC further examine compatibility with multiple residual streams.

\subsection{Experimental Setup}
\label{sec:experimental-setup}

\paragraph{Models and budgets.}
To evaluate recurrent computation at scale, we consider two MoE language models whose attention modules
are implemented with MLA and KDA, respectively. The MLA model has 4.37B active and 69B total parameters,
including 31B over-embedding (OE) parameters~\citep{liu2026scalingemb}; the KDA model has 3.24B active and 51B total parameters, including 10B OE
parameters. Each recurrent model is paired with its corresponding non-recurrent baseline under
iso-parameter and iso-FLOP budgets, and all main experiments use 1T training tokens. Total
parameter counts include shared weights once, whereas active scales follow the unrolled execution
convention, counting each access to the shared Core.

Following the configuration principles in Section~\ref{sec:loop-configuration} ($R=2$,
$\rho \in [25\%, 35\%]$), we configure the MLA and KDA models with unrolled depths
$D_{\mathrm{total}}=28$ ($\rho=28.6\%$) and $D_{\mathrm{total}}=32$ ($\rho=25.0\%$), respectively,
using symmetric Prelude and Coda stages (full layer breakdowns in Appendix~\ref{app:training-details}).
Both configurations preserve their backbone-specific attention computations while incorporating the
same inner-loop state updates and cross-stage residual connection. Reallocating the parameters saved by
weight sharing into extra routed experts per layer restores the removed capacity within the prescribed budgets.

\paragraph{Evaluation.}
We assess model performance on MMLU~\citep{hendrycks2021mmlu}, CMMLU~\citep{li2024cmmlu}, C-Eval~\citep{huang2023ceval}, and BBH~\citep{suzgun2023bbh}.
Recurrent models and their paired baselines share the same evaluation
scope and scoring protocol. Scores are reported as percentages and gains as percentage points (pp).
The aggregate score is the unweighted mean across all four evaluations, and the
aggregate gain is computed relative to the corresponding non-recurrent baseline.

\paragraph{Additional experiments.}
Comparisons with Parcae and SMELT use a common GQA-MoE backbone~\citep{ainslie2023gqa,yang2025qwen3} with approximately 0.8B active and 7.4B total
parameters, trained on 200B tokens from a fixed subset of the public Dolma~3 Mix dataset.
Component ablations use an MLA-MoE recurrent model with approximately 1.3B active and 14.6B total
parameters, including 4.23B OE parameters, evaluated near 100B tokens. The mHC comparison uses the
same token budget and approximately 1.3B active parameters in both models, with 14.5B total parameters
for the non-recurrent baseline and 11.9B for the recurrent model; both include 4.23B OE parameters.
Detailed training protocols, implementation choices, and parameter accounting are provided in
Appendices~\ref{app:training-details}--\ref{app:ablation-mhc}.

\begin{table*}[!t]
    \centering
    \small
    \setlength{\tabcolsep}{9pt}
    \renewcommand{\arraystretch}{1.13}
    \begin{tabular}{lrrrrrr}
        \toprule
        \multirow{2}{*}{Method} & \multicolumn{2}{c}{Parameters (B)}
        & \multicolumn{2}{c}{Dolma Val PPL $\downarrow$} & \multicolumn{2}{c}{LAMBADA PPL $\downarrow$} \\
        \cmidrule(lr){2-3}\cmidrule(lr){4-5}\cmidrule(lr){6-7}
        & Active & Total & 83.89B & 200B & 83.89B & 200B \\
        \midrule
        Baseline & 0.80 & 7.406 & 9.3692 & 8.4948 & 8.8334 & 6.7623 \\
        Parcae   & 0.81 & 7.430 & 9.2175 & 8.3596 & 8.2774 & 6.6870 \\
        SMELT    & 0.81 & 7.361 & 9.2050 & 8.3567 & 8.3895 & 6.4837 \\
        \rowcolor{expOursBlue}
        \textbf{Ours} & 0.81 & 7.427 & \textbf{9.1839} & \textbf{8.3109} & \textbf{8.0987} & \textbf{6.2286} \\
        \bottomrule
    \end{tabular}
    \caption{Comparison on the common GQA-MoE backbone. Active parameter scales are approximate
    unrolled execution counts; total parameters count shared weights once. Both are reported in billions.
    Column groups report PPL at 83.89B and 200B training tokens; lower is better.}
    \label{tab:recurrent-comparison}
\end{table*}

\subsection{Main Results}
\label{sec:main-results}

The unified recurrent design yields consistent gains across both attention architectures under
iso-parameter and iso-FLOP budgets. As shown in Table~\ref{tab:main-results}, the recurrent models outperform their non-recurrent
counterparts on all four evaluations after 1T training tokens. The average score increases from 67.70
to 68.86 with MLA and from 65.42 to 67.71 with KDA, corresponding to improvements of 1.16 and 2.29 pp.

The strongest improvements occur on BBH, where the MLA model advances from 48.09 to 50.95 and the KDA
model from 44.24 to 48.32, yielding gains of 2.86 and 4.08 pp, respectively.

\subsection{Comparison with Existing Recurrent Methods}
\label{sec:recurrent-comparison}

\paragraph{Comparison setup.}
To examine how the proposed information-flow design compares with established recurrent formulations,
we implement our method, Parcae~\citep{prairie2026parcae}, and SMELT~\citep{wang2026smelt} on a common
Qwen3-style GQA-MoE backbone~\citep{yang2025qwen3,ainslie2023gqa}, alongside a non-recurrent reference. Parcae organizes recurrence through
input injection and gated state feedback, whereas SMELT repeats the middle portion of the network and
adjusts internal computation widths and expert capacity to satisfy the parameter and compute constraints. This comparison evaluates
the complete recurrent formulations, with their common-backbone adaptations documented in
Appendix~\ref{app:method-comparison}.

All configurations share a fixed Dolma~3 Mix subset, the Qwen3 Base tokenizer, and an identical training-token
stream, and are trained for 200B tokens. The reference model has approximately 0.80B active and 7.406B
total parameters; the recurrent variants have approximately 0.81B active and 7.4B total parameters,
with individual counts reported in Table~\ref{tab:recurrent-comparison}. Each layer visit activates
eight experts. Our method and Parcae share the same two-pass configuration ($D_{\mathrm{total}}=15$, $R=2$,
$\rho=26.7\%$), expert-pool size, and basic attention setup, while SMELT combines more layer visits
with narrower attention projections and expert intermediate dimensions.

\paragraph{Evaluation protocol.}
We measure language-modeling performance using perplexity (PPL) on held-out Dolma~3 Mix text and
target-word PPL on LAMBADA~\citep{paperno2016lambada}. Dolma validation PPL is the exponential of the validation loss; LAMBADA
is evaluated on the same fixed set of 5,153 test examples in a zero-shot setting. Each metric is
compared across methods under its own scoring protocol.

The proposed framework achieves the lowest perplexity on both evaluations at both training checkpoints
(Table~\ref{tab:recurrent-comparison}). At 200B tokens, Dolma validation PPL reaches 8.3109, compared with
8.3596 for Parcae and 8.3567 for SMELT, while LAMBADA target-word PPL reaches 6.2286, compared with
6.6870 and 6.4837, respectively. Relative to the non-recurrent baseline, these results represent
reductions of 2.16\% on Dolma and 7.89\% on LAMBADA. The same ranking at 83.89B tokens indicates that
the advantage holds at both evaluated stages of training.

\subsection{Component Ablations}
\label{sec:component-ablations}

\paragraph{Ablation setup.}
To examine the contributions of the cross-stage residual connection and inner-loop state updates,
we conduct controlled ablations on an MLA-MoE model with approximately 1.3B active and 14.6B total
parameters, including 4.23B OE parameters. All recurrent variants share the same two-pass Prelude--Core--Coda configuration ($R=2$), expert
capacity, and training setup, and are evaluated near 100B tokens. We remove the cross-stage residual
or replace the inner-loop state-update mechanism while retaining
the remaining components. A depth-matched non-recurrent model with approximately 1.3B active and
14.5B total parameters provides an additional reference. Architecture details and loss aggregation
are specified in Appendix~\ref{app:ablation-mhc}.

\begin{table}[!htbp]
    \centering
    \fontsize{9}{10.5}\selectfont
    \setlength{\tabcolsep}{3pt}
    \renewcommand{\arraystretch}{1.16}
    \begin{tabular}{p{0.47\columnwidth}rr}
        \toprule
        Variant & Val. loss $\downarrow$ & Train loss $\downarrow$ \\
        \midrule
        Non-recurrent baseline & 1.848004 & 1.626291 \\
        \rowcolor{expOursBlue}
        \textbf{Full method} & \textbf{1.801952} & \textbf{1.586823} \\
        No cross-stage residual & 1.832668 & 1.608155 \\
        Additive fusion & 1.804715 & 1.588480 \\
        Concatenation + random & 1.810110 & 1.592462 \\
        \bottomrule
    \end{tabular}
    \caption{Component ablations near 100B tokens. Lower losses are better; validation loss is the
    unweighted mean across the 20 held-out validation sets. \textit{Concatenation + random} applies
    concatenation-based linear fusion from a random initial state. Other components are retained unless specified.}
    \label{tab:component-ablations}
\end{table}

\paragraph{Cross-stage residual connection.}
The cross-stage residual connection provides a direct route from the Prelude representation to the Coda,
complementing the information carried by the recurrent state. Removing this connection increases
validation loss from $1.8020$ to $1.8327$ and training loss from $1.5868$ to $1.6082$
(Table~\ref{tab:component-ablations}). This degradation supports the value of preserving initial
representations through an explicit cross-stage pathway.

\paragraph{Inner-loop state updates.}
To evaluate the proposed inner-loop state-update design, we compare our gated fusion operator (with
direct first-iteration input) against element-wise addition (retaining direct first-iteration input)
and concatenation followed by a linear projection from a random initial state, while keeping the
cross-stage residual connection intact. The two alternatives increase validation loss from $1.8020$
to $1.8047$ and $1.8101$, respectively, with the same ordering in training loss ($1.5868$ vs.\ $1.5885$
and $1.5925$). Although additive fusion remains competitive, the complete method achieves the lowest
training and validation losses across all evaluated configurations, supporting explicit channel-wise
control over state retention and input injection together with the cross-stage residual connection.

\subsection{Integration with mHC}
\label{sec:mhc-evaluation}

Because our looped design explicitly structures inner-loop state updates and cross-stage residual
connectivity, a natural concern is whether its gains overlap with recent architectures that enhance
inter-layer residual information flow, such as mHC. To verify whether recurrent
computation delivers orthogonal gains on top of enhanced residual connectivity, we combine recurrence
with four-stream mHC and compare against a non-recurrent mHC baseline using the same data, sequence
length, global batch size, and approximately 100B training tokens. The recurrent model matches the
unrolled depth of the non-recurrent baseline using a two-pass shared Core ($R=2$; details in
Appendix~\ref{app:ablation-mhc}).

To examine recurrence with the existing mHC configuration, we retain the original expert-pool size
without restoring the parameter capacity removed by sharing. Both models have approximately 1.3B
active parameters, while the recurrent model has 11.9B total parameters compared with 14.5B for the
baseline; both include approximately 4.23B OE parameters. This comparison uses a common training-token
budget while allowing total parameter capacity to differ.

\begin{table}[!htbp]
    \centering
    \fontsize{9}{10.5}\selectfont
    \setlength{\tabcolsep}{3pt}
    \renewcommand{\arraystretch}{1.15}
    \begin{tabular}{lrrrr}
        \toprule
        \multirow{2}{*}{Model} & \multicolumn{2}{c}{Params (B)} & \multirow{2}{*}{Val. loss $\downarrow$} & \multirow{2}{*}{Train loss $\downarrow$} \\
        \cmidrule(lr){2-3}
        & Active & Total & & \\
        \midrule
        mHC & 1.3 & 14.5 & 1.810356 & 1.591382 \\
        \rowcolor{expOursBlue}
        \textbf{mHC + Loop} & 1.3 & 11.9 & \textbf{1.784440} & \textbf{1.572306} \\
        \bottomrule
    \end{tabular}
    \caption{mHC comparison near 100B tokens. Active and total parameter counts are reported in billions
    (shared weights counted once, without parameter restoration). Validation loss is the unweighted mean
    across the 20 held-out validation sets; training loss is averaged over a window near 100B tokens.}
    \label{tab:mhc-results}
\end{table}

The recurrent model improves language modeling while using fewer total parameters
(Table~\ref{tab:mhc-results}). Mean validation loss decreases from $1.8104$ to $1.7844$---with
improvements on 18 of the 20 held-out validation sets---and training loss decreases from $1.5914$ to
$1.5723$. These findings support the compatibility of the proposed information-flow design with mHC
in this setting.

\section{Conclusion}
\label{sec:conclusion}

We presented a framework for scaling looped MoE language models under matched budgets that jointly addresses information flow and parameter allocation.
Inner-loop state updates with gated fusion stabilize iterative refinement, while a cross-stage residual connection preserves a direct residual pathway from the Prelude to the Coda.
Allocating $25\%$--$35\%$ of the unrolled depth to a two-pass shared Core and reallocating the parameters saved by weight sharing into extra routed experts per layer yields consistent gains over iso-parameter and iso-FLOP baselines with MLA and KDA backbones up to 69B parameters, remains effective alongside mHC, and achieves lower perplexity than prior looped methods.
Future work includes extending the analysis to larger recurrence counts and token-adaptive depth, and measuring end-to-end inference efficiency.

\section*{Limitations}

Our study focuses on pre-training looped MoE language models from scratch. First, we do not investigate warm-starting or retrofitting looped recurrence from existing non-recurrent checkpoints. Second, all evaluations are conducted on base models after pre-training without post-training stages such as supervised fine-tuning (SFT) or reinforcement learning (RL), leaving the interaction between looped MoE pre-training and post-training alignment to future work.

\bibliography{custom}

\clearpage
\appendix
\section{Training Details}
\label{app:training-details}

For the 1T-token main experiments, the MLA model uses a $6+8\times2+6$ Prelude--Core--Coda layout
($D_{\mathrm{total}}=28$, $\rho=28.6\%$) and the KDA model uses an $8+8\times2+8$ layout
($D_{\mathrm{total}}=32$, $\rho=25.0\%$).
Table~\ref{tab:kda-training-details} lists the training hyperparameters for the model with KDA attention.
The recurrent-method comparisons and the component/mHC configurations are documented separately in
Appendices~\ref{app:method-comparison} and~\ref{app:ablation-mhc}.

\begin{table}[htbp]
    \centering
    \small
    \setlength{\tabcolsep}{3pt}
    \begin{tabular}{@{}>{\raggedright\arraybackslash}p{0.43\columnwidth}
                        >{\raggedright\arraybackslash}p{0.51\columnwidth}@{}}
        \toprule
        Setting & Value \\
        \midrule
        Sequence length & 8,192 tokens \\
        Micro batch size & 1 \\
        Global batch size & $1{,}024 \to 4{,}096 \to 8{,}192$ \\
        Precision & BF16 \\
        Optimizer & Muon with AdamW parameter groups \\
        Base learning rate & $7.5\times10^{-4}$ \\
        AdamW base rate & $7.5\times10^{-4}$ \\
        Learning-rate schedule & Linear warmup, then constant \\
        Warmup samples & 1,024,000 sequences \\
        Warmup steps & 1,000 \\
        AdamW $(\beta_1,\beta_2)$ & $(0.9,0.95)$ \\
        AdamW $\epsilon$ & $10^{-16}$ \\
        Muon momentum & 0.95 \\
        Muon NS iterations & 5 \\
        Base weight decay & 0.1 \\
        Injection weight decay & 0 \\
        Gradient clipping & 1.0 \\
        Attention / hidden dropout & 0 / 0 \\
        Random seed & 48 \\
        \bottomrule
    \end{tabular}
    \caption{Training hyperparameters for the 1T-token main experiment with KDA attention. Learning rates are base values
    before parameter-group scaling.}
    \label{tab:kda-training-details}
\end{table}

\paragraph{Batch-size and learning-rate schedules.}
The global batch size is 1,024 sequences for the first 6,000 optimizer steps, 4,096 through step 12,000, and 8,192 thereafter.
Learning-rate warmup spans 1,024,000 sequences, corresponding to 1,000 initial optimizer steps and 8,388,608,000 tokens at
the configured sequence length. After warmup, the base learning rate remains constant at $7.5\times10^{-4}$.

\paragraph{Optimizer assignments.}
The implementation uses Muon~\citep{jordan2024muon,liu2025muon} for eligible matrix parameters and AdamW~\citep{loshchilov2019adamw} for the remaining parameter groups. The recurrent
input projection $B$ uses Muon, while the state-retention and input-injection vectors $(\theta, b)$ use AdamW. Recurrent
fusion parameters use zero weight decay in this configuration, and the base weight decay for regularized weights is 0.1.

\paragraph{Training horizon.}
The learning-rate and batch-size schedules are defined over a configured 30T-token training horizon (3,662,109,375 sequences at a
sequence length of 8,192), and all main-experiment evaluations in Section~\ref{sec:experimental-setup} are conducted at the 1T-token checkpoint.

\section{Details of Comparisons with Recurrent Methods}
\label{app:method-comparison}

\subsection{Common Backbone and Parameter Accounting}

All configurations use a Qwen3-style sparse MoE decoder~\citep{yang2025qwen3} with a hidden dimension of 1536, Pre-RMSNorm~\citep{zhang2019rmsnorm},
GQA attention~\citep{ainslie2023gqa}, and SwiGLU experts~\citep{shazeer2020glu}. Each layer visit activates eight experts. The attention head dimension
is 128, the vocabulary contains 151,936 tokens, and the input embedding and output projection are untied.
Every Transformer block contains an MoE module, with no shared experts.

\begin{table*}[t]
    \centering
    \small
    \setlength{\tabcolsep}{6pt}
    \renewcommand{\arraystretch}{1.15}
    \begin{tabular}{llrrrrr}
        \toprule
        Method & Layout & Active / total & Experts/layer & Expert width & Q heads & KV heads \\
        \midrule
        Baseline & 15 independent blocks & 0.80 / 7.406 & 128 & 768 & 16 & 8 \\
        Ours & $2+4\times2+5$ & 0.81 / 7.427 & 176 & 768 & 16 & 8 \\
        Parcae & $2+4\times2+5$ & 0.81 / 7.430 & 176 & 768 & 16 & 8 \\
        SMELT & $4+8\times2+4$ & 0.81 / 7.361 & 192 & 480 & 10 & 5 \\
        \bottomrule
    \end{tabular}
    \caption{Architectures used in the common-backbone comparison. Layouts follow
    Prelude + Core $\times$ iterations + Coda notation. Active and total parameter scales are in billions;
    active scales are approximate unrolled counts. Expert width denotes the intermediate dimension.}
    \label{tab:comparison-configurations}
\end{table*}

Total parameter counts include embeddings, the output projection, attention, experts, routers, normalization,
and parameters introduced by recurrence; shared weights are counted once. Active scales count the parameters
accessed along the unrolled execution path, including the output head and excluding the input embedding.
Thus, a shared Core weight is counted on each visit. These scales do not measure the union of distinct
weights selected by token-dependent routing across iterations.

The non-recurrent reference has approximately 0.569B active backbone parameters, or 0.803B including the
output head. Our method and Parcae use the same layer layout, expert capacity, and basic attention configuration.
The SMELT adaptation uses more layer visits and narrower attention projections and experts, with an enlarged
expert pool. These are the implementations evaluated on our common backbone, rather than the original
papers' complete training configurations.

\subsection{Data and Training Protocol}

All four models use a fixed subset of the public Dolma~3 Mix dataset~\citep{olmo2025olmo3}
(\texttt{allenai/dolma3\_mix-6T}, revision \texttt{689a3ea2\allowbreak d8217e64\allowbreak d73a5058\allowbreak 913fa43a\allowbreak d15e81aa}).
We tokenize the data using the Qwen3 Base tokenizer and partition documents into training, validation,
and test sets using a hash of the original text, preventing exact duplicates from crossing splits.
The models share the processed data snapshot and training-token order, and are trained for 200B tokens.
The selected subset's source composition is recorded in its manifest; we do not assume that its mixture
proportions exactly match those of the complete 6T-token collection.

\begin{table}[t]
    \centering
    \small
    \setlength{\tabcolsep}{3pt}
    \renewcommand{\arraystretch}{1.12}
    \begin{tabular}{>{\raggedright\arraybackslash}p{0.43\columnwidth}>{\raggedright\arraybackslash}p{0.49\columnwidth}}
        \toprule
        Setting & Value \\
        \midrule
        Training tokens & 200B \\
        Sequence length & 4,096 \\
        Global batch size & 2,048 sequences \\
        Precision & BF16 \\
        Optimizer & AdamW \\
        Peak learning rate & $3\times10^{-4}$ \\
        Learning-rate schedule & Linear warmup over approximately 2B tokens, then constant \\
        Adam $(\beta_1,\beta_2)$ & $(0.9,0.95)$ \\
        Adam $\epsilon$ & $10^{-8}$ \\
        Base weight decay & 0.1 \\
        Gradient clipping & 1.0 \\
        Attention / hidden dropout & 0 / 0 \\
        Active experts per visit & 8 \\
        Load balancing & Auxiliary-loss-free expert-bias updates \\
        Expert-bias update rate & $10^{-3}$ \\
        \bottomrule
    \end{tabular}
    \caption{Shared training settings for method comparisons. Method-specific weight-decay
    assignments are described in Appendix~\ref{app:comparison-implementations}.}
    \label{tab:comparison-training}
\end{table}

The routing and auxiliary-loss-free load-balancing rules~\citep{wang2024auxfree} are shared across configurations. Routers use sigmoid scores,
with expert biases affecting expert selection; output weights use the original routing scores.
The auxiliary load-balancing loss coefficient is zero. Shared Core routers and expert biases are reused
across iterations, while expert selection is recomputed from the current hidden states on each visit.

\subsection{Recurrent Implementations}
\label{app:comparison-implementations}

\paragraph{Our method.}
The first iteration feeds the normalized Prelude representation directly into the shared Core.
Subsequent updates mix the recurrent state and fixed input using independently parameterized state-retention
and input-injection coefficients. The Core executes twice, and the original Prelude output is added before
the Coda through the cross-stage residual connection. The input projection $B$, state-retention parameter $\theta$,
and input-injection parameter $b$ are shared across iterations. In this AdamW-trained GQA-MoE setup, fusion parameters receive the main model's
weight decay (0.1), while normalization parameters follow the common normalization-parameter grouping rule.

\paragraph{Parcae.}
We implement random-state initialization, input injection, coupled state retention, and the output projection
on the common backbone, with two fixed iterations and full backpropagation through both.
The initial state follows a truncated normal distribution with standard deviation 0.02 and bounds
$[-0.06,0.06]$. Input and output projections are initialized to the identity. Initial input-injection and
state-retention coefficients are approximately 0.804719 and 0.447214, respectively. Following Parcae's specification, the input/output projections
and state-retention/input-injection vectors use parameter groups without weight decay.

\paragraph{SMELT.}
We repeat the middle eight of 16 independent blocks twice and scale the attention and MoE residual
updates within the recurrent region by $1/2$. The residual-stream dimension remains 1536. The Q/KV
head counts are 10/5, the expert intermediate dimension is 480, and each layer has 192 experts.
Prelude and Coda residual-update coefficients remain 1.

The comparison with Parcae evaluates the complete recurrent configurations, including initialization,
input fusion, the cross-stage residual connection, boundary normalization, and method-specific optimizer grouping.
It is not an isolated ablation of the state-retention parameterization.

\subsection{Evaluation Protocol}

Dolma validation PPL is computed by exponentiating the loss recorded explicitly for the held-out
validation set. LAMBADA reports target-word PPL on a fixed evaluation version containing 5,153 test
examples in the zero-shot setting. All models use the same evaluation data, tokenizer, and scoring
rules; recurrent models retain their two Core iterations at evaluation time. Training budgets in
Table~\ref{tab:recurrent-comparison} refer to effective training tokens.

\section{Ablation and mHC Experimental Details}
\label{app:ablation-mhc}

\subsection{mHC Configuration}

The recurrent mHC model matches the unrolled depth of the non-recurrent mHC baseline using a
Prelude--Core--Coda topology in which the shared Core executes twice ($R=2$), while retaining the
original post-recurrent fusion and normalization boundaries in the Coda.

Both models use four mHC residual streams, hidden dimension 1536, and MLA attention with 16 heads.
The query and key--value low-rank dimensions are 768 and 256, respectively, and the expert intermediate
dimension is 512. The expert-pool configuration is unchanged between the baseline and recurrent model;
the latter does not restore the capacity removed by sharing. Both have approximately 1.3B active
parameters. Total parameter counts are approximately 14.5B and 11.9B, respectively, including
approximately 4.23B OE parameters in each model. The comparison uses the same training-token budget.

\subsection{Component Ablation Configuration}

The ablations use MLA-MoE models without mHC. All recurrent variants share the same two-pass
Prelude--Core--Coda layout ($R=2$), with hidden dimension 1536 and expert intermediate dimension 512.
MLA uses 16 heads with query and key--value low-rank dimensions of 768 and 256. The recurrent
variants incorporate uniform expert-pool restoration to match the parameter budget of the depth-matched
non-recurrent reference (approximately 1.3B active and 14.5B total parameters, including 4.23B OE parameters),
resulting in approximately 1.3B active and 14.6B total parameters (including 4.23B OE parameters).

Across recurrent ablation variants, the backbone and restored expert capacity are held fixed. Removing or replacing the fusion
module changes only a small number of projection and control parameters. The full method and the variant
without the cross-stage residual connection share the same inner-loop state update. Additive fusion replaces the gated operator while retaining direct first-iteration input, and the random-initialization concatenation variant applies concatenation-based linear fusion from a random initial state across both passes.

Parameter scales are estimated from the effective architecture configurations and the repository's
parameter-counting script. Total scales are reported to approximately 0.1B and OE scales to 0.01B.
Active scales include the output head and exclude input embeddings and OE; they use the configured
average expert-activation budget and count shared blocks at each unrolled visit. At approximately
1.3B, the reported precision reflects small additional fusion and connection parameters and does
not imply an exact count of distinct active weights for each routed token.

\subsection{Loss Aggregation}

The mHC and ablation experiments use sequence length 8,192 and global batch size 2,048, corresponding
to 16,777,216 tokens per optimizer step. Training LM loss is averaged over the last 100 complete steps
before 100B tokens: the interval $[5861,5961)$, covering approximately 98.33B--99.99B tokens.
Validation loss is the unweighted mean over our 20 held-out validation sets (spanning general domains, knowledge, math, and code) at step 6,000,
corresponding to approximately 100.66B tokens.

We retrieve and freeze each complete metric series before aggregation. If retries produce multiple
records for the same step, the record with the latest timestamp is retained. All comparisons use
these fixed snapshots and the same aggregation rules. Loss differences, when reported, are the
variant's loss minus the corresponding reference loss.

\end{document}